\documentclass[11pt,a4paper]{article}
\usepackage{authblk}
\usepackage[hyperref]{naaclhlt2019}
\usepackage{times}
\usepackage{latexsym}

\usepackage{url}

\usepackage{multirow}
\usepackage{amsmath}
\usepackage{graphicx}

\aclfinalcopy %
\def\aclpaperid{2139} %

\title{Evaluating Coherence in Dialogue Systems using Entailment}

\author[ ]{Nouha Dziri\Thanks{  Equal Contribution}}
\author[ ]{Ehsan Kamalloo\protect \footnotemark[1]}
\author[ ]{Kory W. Mathewson}
\author[ ]{Osmar Zaiane}
\affil[ ]{Department of Computing Science}
\affil[ ]{University of Alberta}
\affil[ ]{
{\tt \{dziri,kamalloo,korym,zaiane\}@cs.ualberta.ca}}

\date{}

\begin{document}
\maketitle
\begin{abstract}
Evaluating open-domain dialogue systems is difficult due to the diversity of possible correct answers. Automatic metrics such as BLEU correlate weakly with human annotations, resulting in a significant bias across different models and datasets. Some researchers resort to human judgment experimentation for assessing response quality, which is expensive, time consuming, and not scalable. Moreover, judges tend to evaluate a small number of dialogues, meaning that minor differences in evaluation configuration may lead to dissimilar results. In this paper, we present interpretable metrics for evaluating topic coherence by making use of distributed sentence representations. Furthermore, we introduce calculable approximations of human judgment based on conversational coherence by adopting state-of-the-art entailment techniques.  Results show that our metrics can be used as a surrogate for human judgment, making it easy to evaluate dialogue systems on large-scale datasets and allowing an unbiased estimate for the quality of the responses.
\end{abstract}

\section{Introduction}
Recently, we have witnessed a big success in the capability of computers to seemingly understand natural language text and to generate plausible responses to conversations \cite{serban2016building,xing2017topic,sordoni2015neural,li2016persona, serban2017hierarchical, devlin2018bert, radford2018improving}.
A challenging task of building dialogue systems lies in evaluating the quality of their responses. Typically, evaluating goal-oriented dialogue systems is done via human-generated judgment like a task completion test   or user satisfaction score \cite{walker1997paradise, moller2006memo}.
However, the task of evaluating open-ended dialogue systems is not well defined as there is no clear explicit goal for  conversations. Indeed, dialog systems are ultimately created to satisfy the user's need which can be associated with how entertaining and engaging the conversation was. It is  unclear how to define a metric that can account comprehensibly for the semantic meaning of the responses.  Moreover, grasping the underlying meaning of text has always been fraught with difficulties, which are essentially attributed to the complexities and ambiguities in natural language.  
Generally, a good dialogue can be described as an exchange of information that sustain coherence through a train of thoughts and a flow of topics. Therefore, a plausible way to evaluate open-ended dialogue systems is to measure the consistency of the responses. For example, a neural dialogue system can respond to the utterance ``\textit{Do you like animals?}" by ``\textit{Yes, I have three cats}", thereafter replies to ``\textit{How many cats do you have}" by ``\textit{I don't have cats.}". Here, we can notice that the dialogue system failed to provide a coherent answer and instead generated an inconsistent response.

In this work, we characterize the consistency of dialogue systems as a natural language inference (NLI) \cite{dagan2006pascal} problem.  In particular, NLI is focused on recognizing whether a hypothesis is inferred from a premise. In dialogue systems, we cast a generated response as the hypothesis and the conversation history as the premise,  projecting thus the automatic evaluation into an NLI task. In other words, we propose directly calculable approximations of human evaluation grounded on conversational coherence and affordance by using state-of-the-art entailment techniques.
For this purpose, we build a synthesized inference data from conversational corpora. The intuition behind this choice is motivated by the fact that utterances in a human conversation tend to follow a consistent and coherent flow where each utterance can be inferred from the previous interactions. We train the state-of-the-art inference models on our conversational inference data and then the learned models are used to evaluate the coherence in a given conversation. Finally, we fare our proposed evaluation method against existing automated metrics. The results highlight the capability of inference models to automatically evaluate dialogue coherence.  The source code and the dataset are available at \url{https://github.com/nouhadziri/DialogEntailment}

\section{Related Work}

Evaluating open-ended dialogue systems has drawn the attention of several researchers in recent years. Unfortunately, word-overlapping metrics such as BLEU have been shown to correlate weakly with human evaluation, which in turn, introduces bias against certain models \cite{liu2016not}. Many studies have been proposed to improve the quality of automated metrics. In particular, Lowe et al. \cite{lowe2017towards} introduced an automatic evaluation system called ADEM which learns to score responses from an annotated dataset of human responses scores. However, such system is heavily biased towards the training data and  struggles with generalization capabilities on unseen datasets. Further, collecting an annotated gold standard of human judgment is very expensive and thus, ADEM is less flexible and extensible. 
 Venkatesh et al.
\cite{venkatesh2018evaluating} introduced a framework for evaluating the quality of the conversations based on topical diversity, coherence, engagement and conversational depth and showed that these metrics conform with human evaluation. However, a big part of their metrics relies on human labels, which makes the evaluation system not scalable. Recently, Welleck et al. \cite{welleck2018dialogue} investigated the use of NLI models (e.g., ESIM \cite{chen2016enhanced} and InferSent \cite{conneau2017supervised}) to measure consistency in dialogue systems. They built a  Dialogue NLI dataset which consists of sentence pairs labeled as entailment, neutral, or contradiction. The utterances are derived from a two-agent
persona-based dialogue dataset. To annotate the dataset, they used human annotation from Amazon Mechanical Turk.
In this work, we propose a method that employs NLI approaches to detect  coherence in dialogue systems. The proposed evaluation procedure does not require human labels, making progress towards scalable and autonomous evaluation systems.

\section{Natural Language Inference}
Reasoning about the semantic relationship between two utterances is a fundamental part of text understanding. In this setting, we consider inference about entailment as a useful testing bed for the evaluation of coherence in dialogue systems.  
The success of NLI models\footnote{Recent models have achieved high accuracy in Stanford NLI corpus \cite{snli2015emnlp} (90.1\%) and GLUE Benchmark \cite{wang2018glue} (86.7\%)} allows us to frame automated dialogue evaluation as an entailment problem. More specifically, given a conversation history $H$ and a generated response $r$, the goal is to understand whether the premise-hypothesis pair $(H,r)$ is entailing, contradictory, or neutral. 

\subsection{Coherence in Dialogue Systems}
The essence of neural response generation
models is designed by maximizing the likelihood
of the target response given source utterances. Therefore, a dialogue generation task can be formulated as a next utterance prediction problem. In particular, the model predicts a response $u_{i+1}$ given a conversation history $(u_1, ..., u_i)$. One key factor for a successful conversation is having coherence across multiple turns. A machine's response can be considered as incoherent when it contradicts directly its previous utterances or follows an illogical reasoning throughout the whole conversation. 
Inconsistency can be clearly identified when it corresponds to logical discrepancy between two facts.
For example, when you indicate clearly during the conversation that you have cats but when you get asked  ``\textit{How many cats do you have}", you answer by ``\textit{I don't have cats.}". Nevertheless, in general, inconsistency can be less explicitly recognizable as it may describe an error between what the person has said and what she/he truly believes given her/his personality and background information.   \\
To detect dialogue incoherence, we consider two prominent models that have shown promising results in commonsense reasoning: the Enhanced Sequential Inference Model (ESIM) \cite{chen2016enhanced} and Bidirectional Encoder Representations from Transformers (BERT) \cite{devlin2018bert}:

\textbf{ESIM} \cite{chen2016enhanced}: employs a Bi-LSTM model \cite{graves2005framewise} to encode the premise and the hypothesis. Also, it explores the effectiveness of syntax for NLI by encoding syntactic parse trees of premise and hypothesis through Tree-LSTM \cite{zhu2015long}. Then, the input encoding part is followed by a matrix attention layer, a local inference layer, another BiLSTM inference composition layer, and finally a pooling operation before
the output layer. We further boost ESIM with by incorporating contextualized word embeddings, namely ELMo \cite{peters2018deep}, into the inference model.

\textbf{BERT} \cite{devlin2018bert}:  exploits a multi-layer Bidirectional Transformers model \cite{vaswani2017attention} to learn pre-trained universal representations of text using only a plain text corpus from Wikipedia. BERT has achieved state-of-the art results
on various natural language understanding tasks and has been shown to handle strongly long-range dependencies in text. BERT can be fine-tuned to achieve several tasks by solely adding a small layer to the core model. In this work, we adopted BERT to the task of NLI.

Overall, the goal of the above models is to learn a function $G_{NLI}$ that predicts one of three categories (i.e., entailment, contradiction or neutral) given premise-hypothesis pairs.

\section{Inference Corpus for Dialogues}
To train the inference models, we build a synthesized dataset geared toward evaluating consistency in dialogue systems. To this end, the Persona-Chat conversational data \cite{zhang2018personalizing} is used to form the basis of our conversational inference data. The continuity of utterances in human conversation facilitates the use of entailment in the dialogue domain. Typically, when we interact with one another, we tend to reference information from previous utterances to engage with the interlocutor. This is why we build our synthetic inference dataset upon a dialogue corpus. The Persona-Chat corpus is a crowd-sourced dataset where two people converse with each other based on a set of randomly assigned persona. To build an inference corpus, we need to find three different labels (i.e., \emph{entailment}, \emph{contradiction}, and \emph{neutral}). For this purpose, we map an appropriate and on topic response to the \emph{entailment} label. Consequently, the \emph{entailment} instances are derived from the utterances in the conversations. For \emph{contradiction}, grammatically-impaired sentences are constructed by randomly choosing words from the conversation. We also added randomly drawn contradictory instances from the MultiNLI corpus \cite{williams2018broad} to account for meaningful inconsistencies.  Finally, random utterances from other conversations or generic responses such as ``\textit{I don't know}" comprise the \emph{neutral} instances. Following this approach, we build a corpus of 1.1M premise-hypothesis pairs, namely \textbf{InferConvAI}. Table~\ref{tab:DatasetStat} summarizes the statistics of InferConvAI.

 \begin{table}[t]
\centering
\begin{tabular}{ l | c c c }
 \hline
& \textbf{Train} & \textbf{Dev} & \textbf{Test} \\
 \hline
\#entailment & 218.2K & 12.2K & 1.4K \\
\#neutral & 579.5K & 28.0K & 3.1K \\
\#contradiction & 261.9K & 9.8K & 1.1K \\
\hline
\textbf{Total} & 1.1M & 50.2K & 5.6K \\
 \hline
\end{tabular}
\caption{Distribution of labels in the InferConvAI corpus.}
\label{tab:DatasetStat}
\end{table}

\section{Experiments}
In this section, we focus on the task of evaluating the next utterance given the  conversation history. We used the following models to generate responses. These models were trained on the conversational datasets, using optimization, until convergence:

\begin{itemize}
\item Seq2Seq with attention mechanism \cite{bahdanau2014neural}: predicts the next response given the previous utterance using an encoder-decoder model.
\setlength{\abovedisplayskip}{1pt}
 \item HRED \cite{serban2016building}: extends the Seq2Seq model by adding a context-RNN layer that accounts for contextual information.
  \setlength{\belowdisplayskip}{1pt}
  \item TA-Seq2Seq \cite{xing2017topic}: extends the Seq2Seq model by biasing the overall distribution towards leveraging topic words in the response.
  \item THRED \cite{dziri2018augmenting}: builds upon TA-Seq2Seq model by levering topic words in the response in a multi-turn dialogue system.
  \setlength{\belowdisplayskip}{1pt}

\end{itemize}
 The training was conducted on two datasets: OpenSubtitles \cite{tiedemann2012parallel} and Reddit \cite{dziri2018augmenting}. Due to lack of resources, we randomly sampled 6M dialogues as training data from each dataset, 700K dialogues as development data, and 40K dialogues as test data. Each dialogue corresponds to three turn exchanges.   To evaluate accurately the quality of the generated responses, we recruited five  native English speakers. The judges annotated 150 dialogues from Reddit and 150 dialogues from OpenSubtitles. All subjects have informed consent as required from the Ethics Review Board at the University of Alberta.
 Due to lack of space, we will omit an exhaustive description of the human evaluation process and refer readers to \cite{dziri2018augmenting} as we conducted the same evaluation procedure.

\begin{table}[t]
\centering
\begin{tabular}{ l | c c }
 \hline
 \textbf{Method} & \textbf{Reddit} & \textbf{OpenSubtitles} \\
 \hline
ESIM + ELMo & 0.526 & 0.455 \\
BERT & \textbf{0.553} & \textbf{0.498} \\
 \hline
\end{tabular}
\caption{Accuracy of inference models on InferConvAI.}
\label{tab:Entailment}
\end{table}

\begin{figure}[t]
\centering
\includegraphics[width=.98\linewidth]{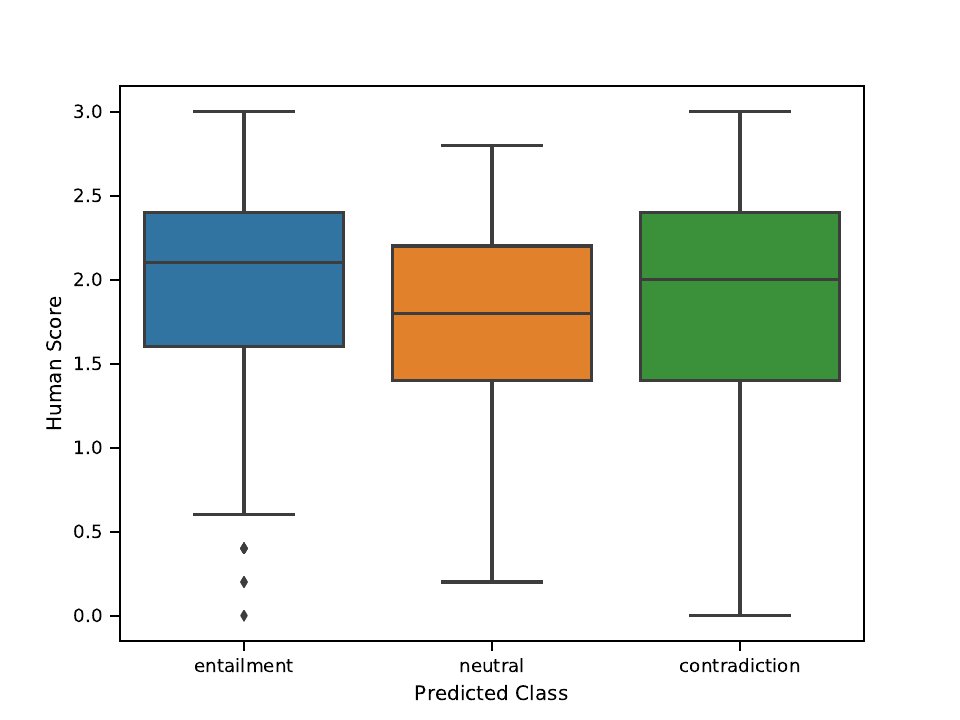}
\caption{BERT predictions for each class vs. human scores. The labels in the horizontal axis are (from left to right): entailment, neutral, contradiction.}
\label{fig:InferenceVsHumanScore}
\end{figure}

\subsection{NLI in Dialogues}
In this section, we evaluate the performance of the state-of-the-art entailment models on predicting a score for the generated utterances. In particular, the conversation history $H$ is treated as a premise, whereas the generated response $r$ acts as a hypothesis. We pick two state-of-the-art NLI models (i.e., ESIM \cite{chen2016enhanced} and BERT \cite{devlin2018bert}). These models were trained on the InferConvAI dataset. During evaluation, we use our test dialogue corpus from Reddit and OpenSubtitles, in which the majority vote of the 4-scale human rating constitutes the labels. The results are illustrated in Table~\ref{tab:Entailment}. Both models reach reasonable performance in this setting, while BERT outperforms ESIM. Note that this experiment examines the generalization capabilities of these inference models as the test datasets are drawn from an entirely different distribution than the training corpus. Figure~\ref{fig:InferenceVsHumanScore} illustrates the performance of BERT for each class with respect to the human scores. The test utterances that are predicted as \emph{entailment} tend to be rated higher than other utterances, exhibiting that the entailment models correlate quite well with what humans perceive as a coherent response. Another observation is that the inference models often classify acontextual and off-topic responses as \emph{neutral} and the annotators typically dislike these types of responses. This contributes to the lower scores of \emph{neutral}-detected responses compared to responses predicted as \emph{contradiction}.

\begin{figure*}[!tb]
\centering
\minipage{.33\textwidth}
\includegraphics[width=\linewidth]{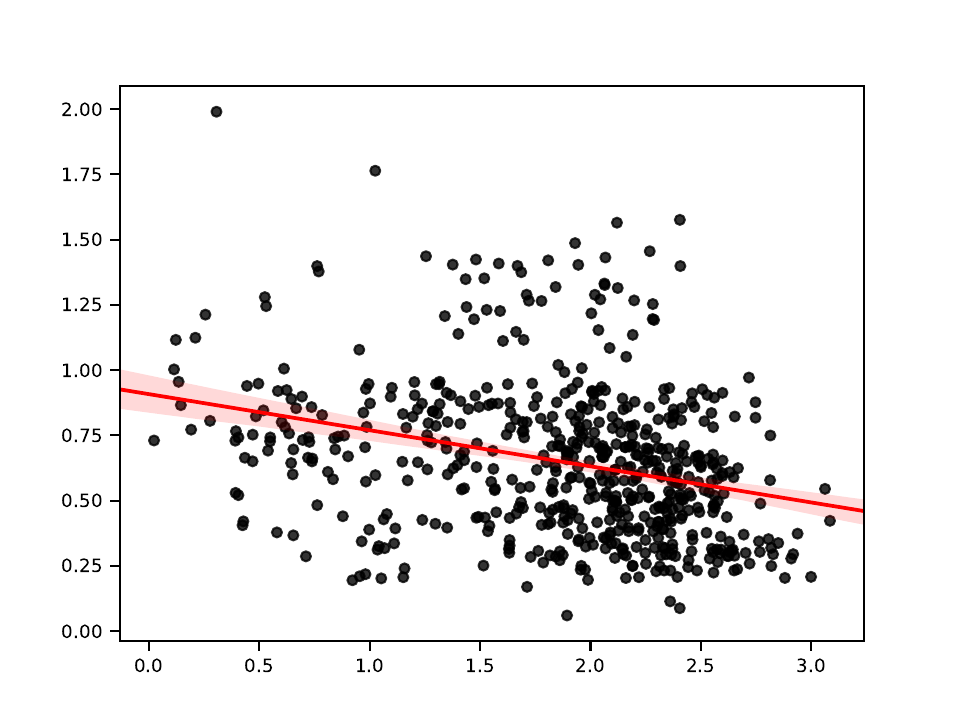}
\endminipage
\minipage{.33\textwidth}
\includegraphics[width=\linewidth]{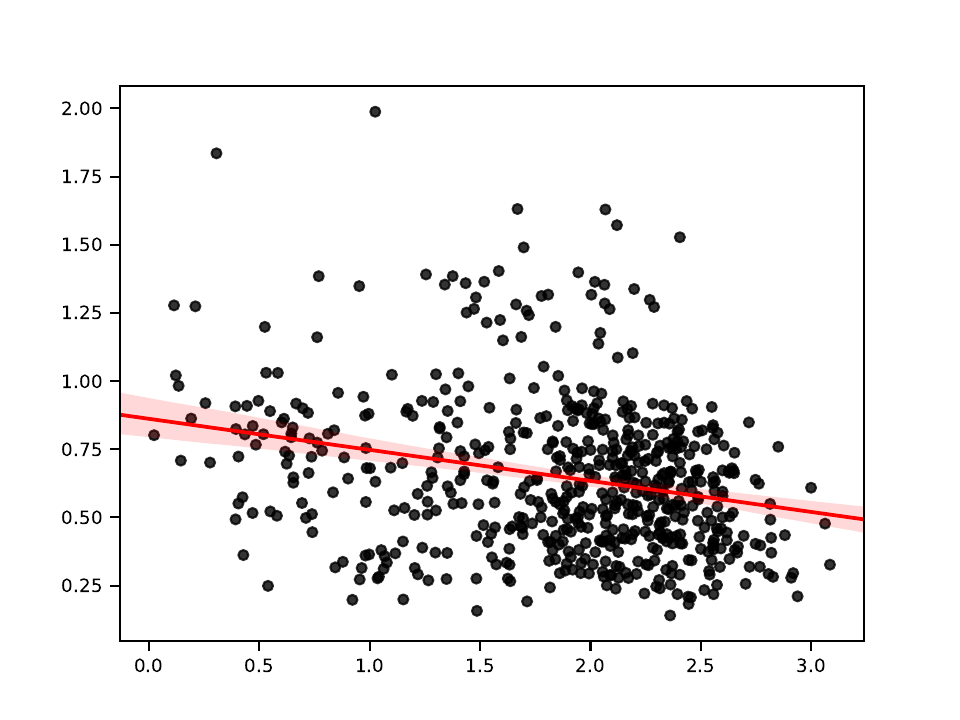}
\endminipage
\minipage{.33\textwidth}
\includegraphics[width=\linewidth]{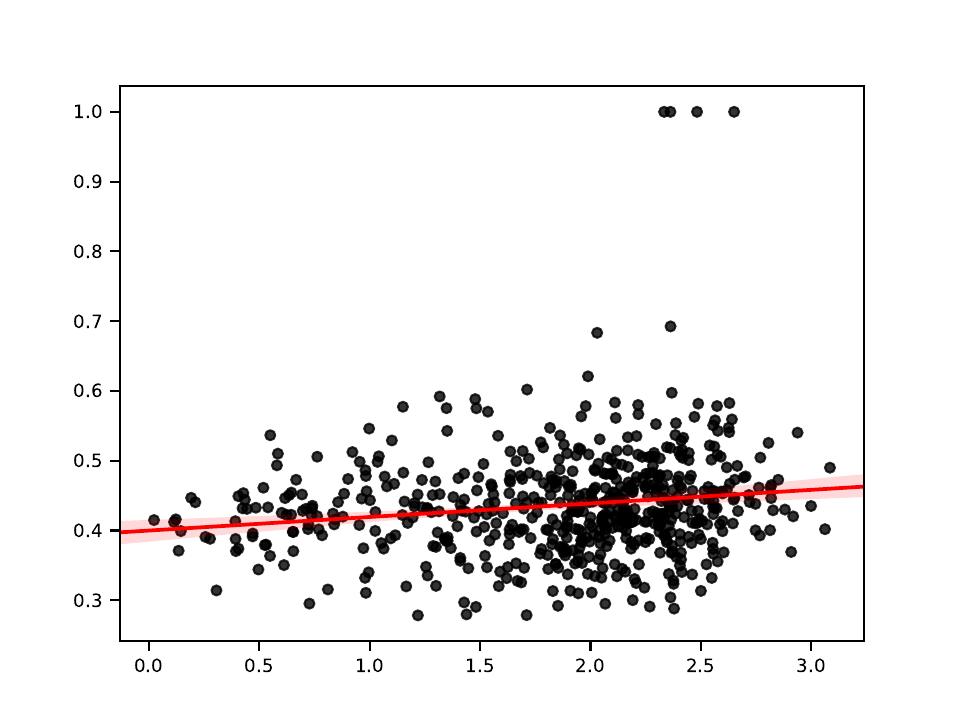}
\endminipage
\caption{{\small Scatter plots illustrating correlation between human judgment and the automated metrics on the Reddit test dataset. In order to better visualize the density of the points, we added stochastic noise generated by Gaussian distribution $\mathcal{N}(0, 0.1)$ to the human ratings (i.e., horizontal axis) at the cost of lowering correlation, as done in \cite{lowe2017towards}. From left to right: SS$_{\mathrm{USE}}$ w.r.t. the second most recent utterance ($H_{-2}$), SS$_\mathrm{USE}$ w.r.t. the most recent utterance ($H_{-1}$), and Extrema$_\mathrm{BERT}$}}
\label{fig:correlation_scatter}
\end{figure*}

\begin{table}[t]
\centering
\begin{tabular}{ l | c c }
 \hline
 \multirow{2}{*}{\textbf{Method}} & \multicolumn{2}{c}{\textbf{Pearson}} \\
 & {\small Reddit} & {\small OpenSubtitles} \\
 \hline
SS$(H_{-2})_\mathrm{BERT}$ & -0.204 & -0.290 \\
SS$(H_{-2})_\mathrm{ELMo}$ & -0.146 & \underline{-0.365} \\
SS$(H_{-2})_\mathrm{USE}$ & \underline{-0.248} & -0.314 \\
SS$(H_{-1})_\mathrm{BERT}$ & -0.214 & -0.337 \\
SS$(H_{-1})_\mathrm{ELMo}$ & -0.178 & \textbf{-0.404} \\
SS$(H_{-1})_\mathrm{USE}$ & \textbf{-0.287} & -0.320 \\
A$_\mathrm{BERT}$ & 0.135 & 0.131 \\
A$_\mathrm{ELMo}$ & 0.085 & 0.162 \\
A$_\mathrm{word2vec}$ & 0.037 & 0.196 \\
G$_\mathrm{BERT}$ & 0.208 & 0.132 \\
G$_\mathrm{ELMo}$ & 0.037 & 0.072 \\
G$_\mathrm{word2vec}$ & -0.033 & 0.015 \\
E$_\mathrm{BERT}$ & 0.162 & 0.144 \\
E$_\mathrm{ELMo}$ & 0.035 & 0.116 \\
E$_\mathrm{word2vec}$ & -0.065 & 0.118 \\
 \hline
\end{tabular}
\caption{The Pearson Correlation between different metrics and human judgments with $p$-value $< 0.001$. The semantic similarity (SS) metric is measured with respect to the most recent utterance $H_{-1}$ and the most recent two utterances $H_{-2}$ in the conversation history. We adopt different  embedding  algorithms to compute the word vectors:  ELMo \cite{peters2018deep}, BERT \cite{devlin2018bert}, word2vec \cite{mikolov2013distributed} and Universal Sentence Encoder (USE) \cite{cer2018universal}.}
\label{tab:Pearson}
\end{table}

\subsection{Automated Metrics}

\subsubsection{Word-level Metrics}
We consider as evaluation metrics baselines three textual similarity metrics \cite{liu2016not} based on word embeddings: Average (A), Greedy (G), and Extrema (E). These word-level embedding metrics have been proven to correlate with human judgment marginally better than other world-overlap metrics (e.g., BLEU, ROUGE and METEOR) \cite{liu2016not}. One critical flaw of these embedding metrics is that they assume that each word is independent of the other words in the sentence. Further, the sentence is treated as a bag-of-words, disregarding words order and dependencies that are known to be substantial for understanding the semantic of a sentence.
 The correlation of these metrics with human judgment is showcased in Table~\ref{tab:Pearson}.
 We can notice that the three metrics A, G and E correlate weakly with human judgment in both datasets, demonstrating the need for a well-designed automated metric that provides an accurate evaluation of dialogues.

\subsubsection{Semantic Similarity}

The Semantic Similarity (SS) metric  was suggested by \cite{dziri2018augmenting}. It measures the distance between the generated response and the utterances in the conversation history.  The intuition of this metric revolves around capturing good and consistent responses by showing whether the machine-generated responses maintain  the topic of the conversation. 
In this project, we measured SS with respect to two different utterances, the conversation history $H$ and the most recent utterance $H_{-1}$. The conversation history is formed by concatenating the two most recent utterances. We report the Pearson coefficient of this metric with human judgment in Table~\ref{tab:Pearson}. The SS metric is expected to have a negative correlation as the higher human ratings correspond to the lower semantic distance. 
 The results demonstrate that SS metrics correlate better than word-level metrics as they make use of word interactions to represent utterances. Moreover, the Universal Sentence Encoder (USE) \cite{cer2018universal} model performs better on Reddit, whereas the ELMo embeddings achieve higher correlation on OpenSubtitles. This arguably underlines that deep contextualized word representations can manage better complex characteristics of natural language (e.g., syntax and semantics). The SS metric, which requires no pre-training, reaches a Pearson correlation of -0.404 with respect to the most recent utterance on OpenSubtitles. Such correlation can be compared with a correlation of 0.436 achieved by ADEM \cite{lowe2017towards} which required large amounts of training data and computation.
Moreover, in order to investigate whether the results in Table~\ref{tab:Pearson} are in line with human evaluation, we visualized the correlation between the human ratings and SS metric as scatter plots in Figure~\ref{fig:correlation_scatter}.

\section{Conclusion}
Evaluating dialogue systems has been heavily investigated, but researchers are still on the quest for a strong and reliable metric that highly conforms with human judgment. Existing automated metrics show poor correlation with human annotations. In this paper, we present a novel paradigm for evaluating the coherence of dialogue systems by using state-of-the-art entailment techniques. We aim at building a system that does not require human annotation, which in turn, can lead to a scalable evaluation approach. 
While our results illustrate that the proposed approach correlates reasonably with human judgment and provide an unbiased estimate for the response quality, we believe that there is still room for improvement. For instance, measuring the engagingness of the conversation would be helpful in improving evaluating different dialogue strategies.

\bibliography{naaclhlt2019}

\begin{thebibliography}{29}
\expandafter\ifx\csname natexlab\endcsname\relax\def\natexlab#1{#1}\fi

\bibitem[{{Bahdanau} et~al.(2015){Bahdanau}, {Cho}, and
  {Bengio}}]{bahdanau2014neural}
Dzmitry {Bahdanau}, Kyunghyun {Cho}, and Yoshua {Bengio}. 2015.
\newblock Neural machine translation by jointly learning to align and
  translate.
\newblock \emph{international conference on learning representations}.

\bibitem[{Bowman et~al.(2015)Bowman, Angeli, Potts, and
  Manning}]{snli2015emnlp}
Samuel~R. Bowman, Gabor Angeli, Christopher Potts, and Christopher~D. Manning.
  2015.
\newblock A large annotated corpus for learning natural language inference.
\newblock In \emph{Proceedings of the 2015 Conference on Empirical Methods in
  Natural Language Processing (EMNLP)}. Association for Computational
  Linguistics.

\bibitem[{Cer et~al.(2018)Cer, Yang, Kong, Hua, Limtiaco, John, Constant,
  Guajardo-Cespedes, Yuan, Tar et~al.}]{cer2018universal}
Daniel Cer, Yinfei Yang, Sheng-yi Kong, Nan Hua, Nicole Limtiaco, Rhomni~St
  John, Noah Constant, Mario Guajardo-Cespedes, Steve Yuan, Chris Tar, et~al.
  2018.
\newblock Universal sentence encoder.
\newblock \emph{arXiv preprint arXiv:1803.11175}.

\bibitem[{Chen et~al.(2016)Chen, Zhu, Ling, Wei, Jiang, and
  Inkpen}]{chen2016enhanced}
Qian Chen, Xiaodan Zhu, Zhenhua Ling, Si~Wei, Hui Jiang, and Diana Inkpen.
  2016.
\newblock Enhanced lstm for natural language inference.
\newblock \emph{arXiv preprint arXiv:1609.06038}.

\bibitem[{Conneau et~al.(2017)Conneau, Kiela, Schwenk, Barrault, and
  Bordes}]{conneau2017supervised}
Alexis Conneau, Douwe Kiela, Holger Schwenk, Loic Barrault, and Antoine Bordes.
  2017.
\newblock Supervised learning of universal sentence representations from
  natural language inference data.
\newblock \emph{arXiv preprint arXiv:1705.02364}.

\bibitem[{Dagan et~al.(2006)Dagan, Glickman, and Magnini}]{dagan2006pascal}
Ido Dagan, Oren Glickman, and Bernardo Magnini. 2006.
\newblock The pascal recognising textual entailment challenge.
\newblock In \emph{Machine Learning Challenges. Evaluating Predictive
  Uncertainty, Visual Object Classification, and Recognising Tectual
  Entailment}, pages 177--190, Berlin, Heidelberg. Springer Berlin Heidelberg.

\bibitem[{Devlin et~al.(2018)Devlin, Chang, Lee, and
  Toutanova}]{devlin2018bert}
Jacob Devlin, Ming-Wei Chang, Kenton Lee, and Kristina Toutanova. 2018.
\newblock Bert: Pre-training of deep bidirectional transformers for language
  understanding.
\newblock \emph{arXiv preprint arXiv:1810.04805}.

\bibitem[{Dziri et~al.(2018)Dziri, Kamalloo, Mathewson, and
  Zaiane}]{dziri2018augmenting}
Nouha Dziri, Ehsan Kamalloo, Kory~W Mathewson, and Osmar Zaiane. 2018.
\newblock Augmenting neural response generation with context-aware topical
  attention.
\newblock \emph{arXiv preprint arXiv:1811.01063}.

\bibitem[{Graves and Schmidhuber(2005)}]{graves2005framewise}
Alex Graves and J{\"u}rgen Schmidhuber. 2005.
\newblock Framewise phoneme classification with bidirectional lstm and other
  neural network architectures.
\newblock \emph{Neural Networks}, 18(5-6):602--610.

\bibitem[{{Li} et~al.(2016){Li}, {Galley}, {Brockett}, {Spithourakis}, {Gao},
  and {Dolan}}]{li2016persona}
Jiwei {Li}, Michel {Galley}, Chris {Brockett}, Georgios~P. {Spithourakis},
  Jianfeng {Gao}, and Bill {Dolan}. 2016.
\newblock A persona-based neural conversation model.
\newblock In \emph{Proceedings of the 54th Annual Meeting of the Association
  for Computational Linguistics (Volume 1: Long Papers)}, volume~1, pages
  994--1003.

\bibitem[{{Liu} et~al.(2016){Liu}, {Lowe}, {Serban}, {Noseworthy}, {Charlin},
  and {Pineau}}]{liu2016not}
Chia-Wei {Liu}, Ryan {Lowe}, Iulian~Vlad {Serban}, Michael {Noseworthy},
  Laurent {Charlin}, and Joelle {Pineau}. 2016.
\newblock How not to evaluate your dialogue system: An empirical study of
  unsupervised evaluation metrics for dialogue response generation.
\newblock In \emph{Proceedings of the 2016 Conference on Empirical Methods in
  Natural Language Processing}, pages 2122--2132.

\bibitem[{{Lowe} et~al.(2017){Lowe}, {Noseworthy}, {Serban},
  {Angelard-Gontier}, {Bengio}, and {Pineau}}]{lowe2017towards}
Ryan {Lowe}, Michael {Noseworthy}, Iulian~Vlad {Serban}, Nicolas
  {Angelard-Gontier}, Yoshua {Bengio}, and Joelle {Pineau}. 2017.
\newblock Towards an automatic turing test: Learning to evaluate dialogue
  responses.
\newblock In \emph{Proceedings of the 55th Annual Meeting of the Association
  for Computational Linguistics (Volume 1: Long Papers)}, volume~1, pages
  1116--1126.

\bibitem[{Mikolov et~al.(2013)Mikolov, Sutskever, Chen, Corrado, and
  Dean}]{mikolov2013distributed}
Tomas Mikolov, Ilya Sutskever, Kai Chen, Greg~S Corrado, and Jeff Dean. 2013.
\newblock Distributed representations of words and phrases and their
  compositionality.
\newblock In \emph{Advances in neural information processing systems}, pages
  3111--3119.

\bibitem[{M{\"o}ller et~al.(2006)M{\"o}ller, Englert, Engelbrecht, Hafner,
  Jameson, Oulasvirta, Raake, and Reithinger}]{moller2006memo}
Sebastian M{\"o}ller, Roman Englert, Klaus Engelbrecht, Verena Hafner, Anthony
  Jameson, Antti Oulasvirta, Alexander Raake, and Norbert Reithinger. 2006.
\newblock Memo: towards automatic usability evaluation of spoken dialogue
  services by user error simulations.
\newblock In \emph{Ninth International Conference on Spoken Language
  Processing}.

\bibitem[{Peters et~al.(2018)Peters, Neumann, Iyyer, Gardner, Clark, Lee, and
  Zettlemoyer}]{peters2018deep}
Matthew~E Peters, Mark Neumann, Mohit Iyyer, Matt Gardner, Christopher Clark,
  Kenton Lee, and Luke Zettlemoyer. 2018.
\newblock Deep contextualized word representations.
\newblock \emph{arXiv preprint arXiv:1802.05365}.

\bibitem[{Radford et~al.(2018)Radford, Narasimhan, Salimans, and
  Sutskever}]{radford2018improving}
Alec Radford, Karthik Narasimhan, Tim Salimans, and Ilya Sutskever. 2018.
\newblock Improving language understanding by generative pre-training.
\newblock \emph{URL https://s3-us-west-2. amazonaws.
  com/openai-assets/research-covers/languageunsupervised/language understanding
  paper. pdf}.

\bibitem[{Serban et~al.(2016)Serban, Sordoni, Bengio, Courville, and
  Pineau}]{serban2016building}
Iulian~Vlad Serban, Alessandro Sordoni, Yoshua Bengio, Aaron~C Courville, and
  Joelle Pineau. 2016.
\newblock Building end-to-end dialogue systems using generative hierarchical
  neural network models.
\newblock In \emph{AAAI}, pages 3776--3784.

\bibitem[{Serban et~al.(2017)Serban, Sordoni, Lowe, Charlin, Pineau, Courville,
  and Bengio}]{serban2017hierarchical}
Iulian~Vlad Serban, Alessandro Sordoni, Ryan Lowe, Laurent Charlin, Joelle
  Pineau, Aaron~C Courville, and Yoshua Bengio. 2017.
\newblock A hierarchical latent variable encoder-decoder model for generating
  dialogues.
\newblock In \emph{AAAI}, pages 3295--3301.

\bibitem[{{Sordoni} et~al.(2015){Sordoni}, {Galley}, {Auli}, {Brockett}, {Ji},
  {Mitchell}, {Nie}, {Gao}, and {Dolan}}]{sordoni2015neural}
Alessandro {Sordoni}, Michel {Galley}, Michael {Auli}, Chris {Brockett},
  Yangfeng {Ji}, Margaret {Mitchell}, Jian-Yun {Nie}, Jianfeng {Gao}, and Bill
  {Dolan}. 2015.
\newblock A neural network approach to context-sensitive generation of
  conversational responses.
\newblock In \emph{Proceedings of the 2015 Conference of the North American
  Chapter of the Association for Computational Linguistics: Human Language
  Technologies}, pages 196--205.

\bibitem[{Tiedemann(2012)}]{tiedemann2012parallel}
J{\"o}rg Tiedemann. 2012.
\newblock Parallel data, tools and interfaces in opus.
\newblock In \emph{LREC}, volume 2012, pages 2214--2218.

\bibitem[{Vaswani et~al.(2017)Vaswani, Shazeer, Parmar, Uszkoreit, Jones,
  Gomez, Kaiser, and Polosukhin}]{vaswani2017attention}
Ashish Vaswani, Noam Shazeer, Niki Parmar, Jakob Uszkoreit, Llion Jones,
  Aidan~N Gomez, {\L}ukasz Kaiser, and Illia Polosukhin. 2017.
\newblock Attention is all you need.
\newblock In \emph{Advances in Neural Information Processing Systems}, pages
  5998--6008.

\bibitem[{{Venkatesh} et~al.(2018){Venkatesh}, {Khatri}, {Ram}, {Guo},
  {Gabriel}, {Nagar}, {Prasad}, {Cheng}, {Hedayatnia}, {Metallinou}, {Goel},
  {Yang}, and {Raju}}]{venkatesh2018evaluating}
Anu {Venkatesh}, Chandra {Khatri}, Ashwin {Ram}, Fenfei {Guo}, Raefer
  {Gabriel}, Ashish {Nagar}, Rohit {Prasad}, Ming {Cheng}, Behnam {Hedayatnia},
  Angeliki {Metallinou}, Rahul {Goel}, Shaohua {Yang}, and Anirudh {Raju}.
  2018.
\newblock On evaluating and comparing conversational agents.
\newblock \emph{arXiv preprint arXiv:1801.03625}.

\bibitem[{Walker et~al.(1997)Walker, Litman, Kamm, and
  Abella}]{walker1997paradise}
Marilyn~A Walker, Diane~J Litman, Candace~A Kamm, and Alicia Abella. 1997.
\newblock Paradise: A framework for evaluating spoken dialogue agents.
\newblock In \emph{Proceedings of the eighth conference on European chapter of
  the Association for Computational Linguistics}, pages 271--280. Association
  for Computational Linguistics.

\bibitem[{Wang et~al.(2018)Wang, Singh, Michael, Hill, Levy, and
  Bowman}]{wang2018glue}
Alex Wang, Amapreet Singh, Julian Michael, Felix Hill, Omer Levy, and Samuel~R
  Bowman. 2018.
\newblock Glue: A multi-task benchmark and analysis platform for natural
  language understanding.
\newblock \emph{arXiv preprint arXiv:1804.07461}.

\bibitem[{Welleck et~al.(2018)Welleck, Weston, Szlam, and
  Cho}]{welleck2018dialogue}
Sean Welleck, Jason Weston, Arthur Szlam, and Kyunghyun Cho. 2018.
\newblock Dialogue natural language inference.
\newblock \emph{arXiv preprint arXiv:1811.00671}.

\bibitem[{Williams et~al.(2018)Williams, Nangia, and
  Bowman}]{williams2018broad}
Adina Williams, Nikita Nangia, and Samuel Bowman. 2018.
\newblock A broad-coverage challenge corpus for sentence understanding through
  inference.
\newblock In \emph{Proceedings of the 2018 Conference of the North American
  Chapter of the Association for Computational Linguistics: Human Language
  Technologies, Volume 1 (Long Papers)}, volume~1, pages 1112--1122.

\bibitem[{Xing et~al.(2017)Xing, Wu, Wu, Liu, Huang, Zhou, and
  Ma}]{xing2017topic}
Chen Xing, Wei Wu, Yu~Wu, Jie Liu, Yalou Huang, Ming Zhou, and Wei-Ying Ma.
  2017.
\newblock Topic aware neural response generatio.
\newblock In \emph{AAAI}, pages 3351--3357.

\bibitem[{Zhang et~al.(2018)Zhang, Dinan, Urbanek, Szlam, Kiela, and
  Weston}]{zhang2018personalizing}
Saizheng Zhang, Emily Dinan, Jack Urbanek, Arthur Szlam, Douwe Kiela, and Jason
  Weston. 2018.
\newblock Personalizing dialogue agents: I have a dog, do you have pets too?
\newblock \emph{arXiv preprint arXiv:1801.07243}.

\bibitem[{Zhu et~al.(2015)Zhu, Sobihani, and Guo}]{zhu2015long}
Xiaodan Zhu, Parinaz Sobihani, and Hongyu Guo. 2015.
\newblock Long short-term memory over recursive structures.
\newblock In \emph{International Conference on Machine Learning}, pages
  1604--1612.

\end{thebibliography}
\bibliographystyle{acl_natbib}

\end{document}